\documentclass[10pt,twocolumn,letterpaper]{article}

\usepackage{iccv}
\usepackage{times}
\usepackage{epsfig}
\usepackage{graphicx}
\usepackage{amsmath}
\usepackage{amssymb}
\usepackage{multirow}
\usepackage{booktabs}
\usepackage{fancyhdr}
\usepackage{authblk}

\usepackage[breaklinks=true,bookmarks=false]{hyperref}

\iccvfinalcopy 

\def\iccvPaperID{****} 

\ificcvfinal\fi

\begin{document}

\title{Vatex Video Captioning Challenge 2020: \\Multi-View Features and Hybrid Reward Strategies for Video Captioning}

\author[$^{1}$ ]{
	 Xinxin Zhu  $^*$
}
\author[$^{1}$]{\ \ \ 
	\; Longteng Guo $^*$ 
}
\author[$^{2}$]{\ \ \ 
	\; Peng Yao \thanks{The authors contributed equally. Peng Yao was an intern at Institute of Automation, Chinese Academy of Sciences during this work.}   } 
\author[$^{3}$]{\ \ \ 
	\;Shichen Lu  
}
\author[$^{1}$]{\ \ \ 
	\;Wei Liu
}
\author[$^{1}$]{\ \ \ 
	\; Jing Liu\thanks{Corresponding Author}  
}
\affil[ ]{$^1$National Laboratory of Pattern Recognition, Institute of Automation, Chinese Academy of Sciences; $^2$University of Science and Technology Beijing; $^3$Wuhan University.}
\affil[ ]{ 
	\tt\small \{xinxin.zhu,longteng.guo,jliu\}@nlpr.ia.ac.cn, pengyao@xs.ustb.edu.cn,sclu@whu.edu.cn, liuwei2019@ia.ac.cn}

\renewcommand\Authsep{ } 
\renewcommand\Authands{ }

\maketitle
\ificcvfinal\thispagestyle{empty}\fi

\begin{abstract}
This report describes our solution for the VATEX Captioning Challenge 2020, which requires generating descriptions for the videos in both English and Chinese languages. We identified three crucial factors that improve the performance, namely: multi-view features, hybrid reward, and diverse ensemble. Based on our method of VATEX 2019 challenge, we achieved significant improvements this year with more advanced model architectures, combination of appearance and motion features, and careful hyper-parameters tuning. Our method achieves the first place on both the Chinese and English video captioning tracks, respectively.	
\end{abstract}

\section{Introduction}
Video captioning aims at describing the video contents with natural language. 
The fundamental challenge of this task is to accurately recognize the activities in a video clip, 
and depict them with high-quality, diverse captions. 
To promote the progress in video captioning, the VATEX Captioning Challenge is held. 
The training and test sets in this challenge are from the recently proposed largest multilingual dataset, i.e. VATEX \cite{wang2019vatex}. 
VATEX contains videos covering 600 human activities, with each video paired with 10 English and 10
Chinese diverse captions. 

\begin{figure}[t]
	\begin{center}
		\includegraphics[width=0.99\linewidth]{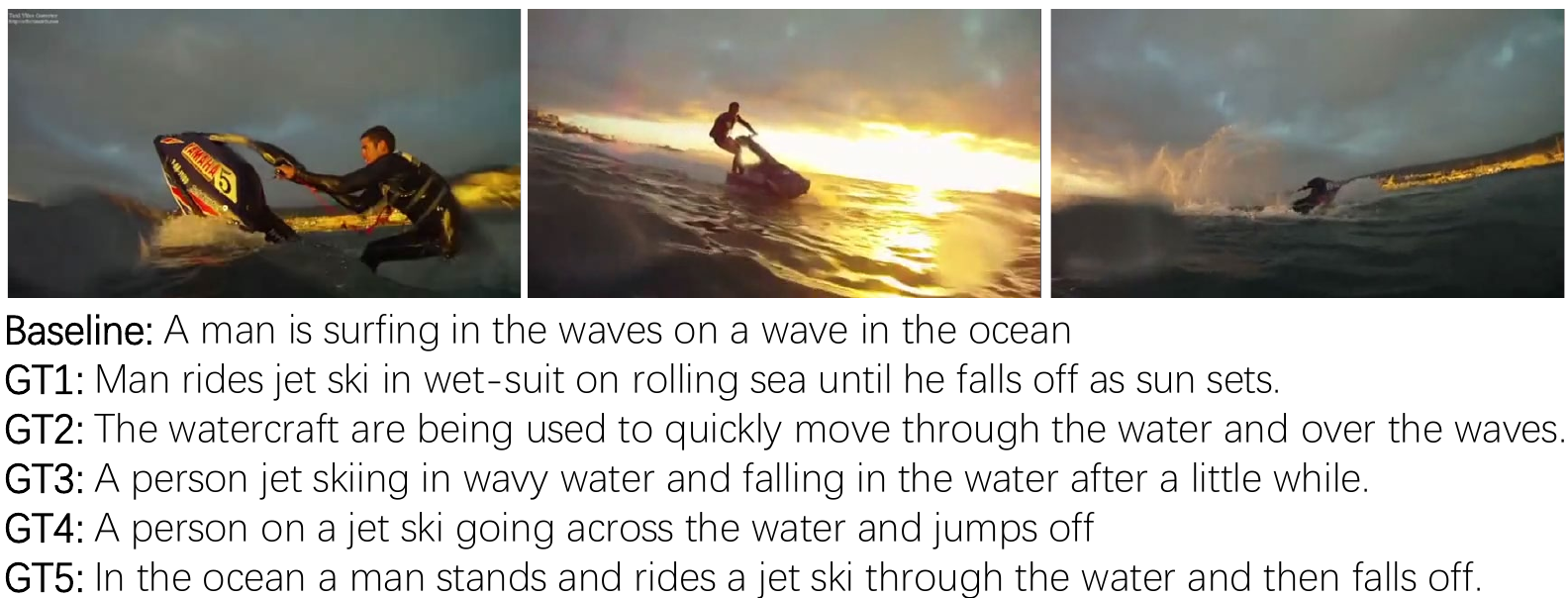}
	\end{center}
	\caption{Example of captions from our baseline model and the ground-truth captions.}
	\label{ff}
\end{figure}

The biggest challenge of this dataset is the large variety of video content, 
which is often very challenging to recognize. 
An example is shown in Figure~\ref{ff}, where we show a video, the corresponding English caption generated by our baseline model and the ground-truth captions.  
Our baseline video captioning model mistakes the activity, ``jet skiing", in the video as ``surfing" and the man ``falls off" is ignored because this action too fast to be noticed. 
Another challenge in this task is the vast diversity of the captions. 
Still, we take Figure \ref{ff} for example. 
The ground-truth captions for this specific video differ a lot either in the caption content or in the caption length. 

In this work, we propose multi-view features and hybrid reward methods to address the above two challenges respectively. 
Multi-view features extracted by different backbone models aim to provide more comprehensive and discriminative video representation, including appearance and motion features.
The hybrid reward method directly optimizes a linear combination of CIDEr, METEOR, ROUGE and BLEU scores during the reinforcement learning stage, which we found performs significantly better than solely optimizing one of them. 
Furthermore, by using diverse ensemble of two state-of-the-art caption frameworks and different video features as inputs, we achieve CIDEr scores of 59.5 and 81.4 on the test-Chinese and test-English splits of the VATEX dataset respectively and win the first place on both tracks.

\begin{figure*}[t]
	\begin{center}
		\includegraphics[width=0.4\linewidth]{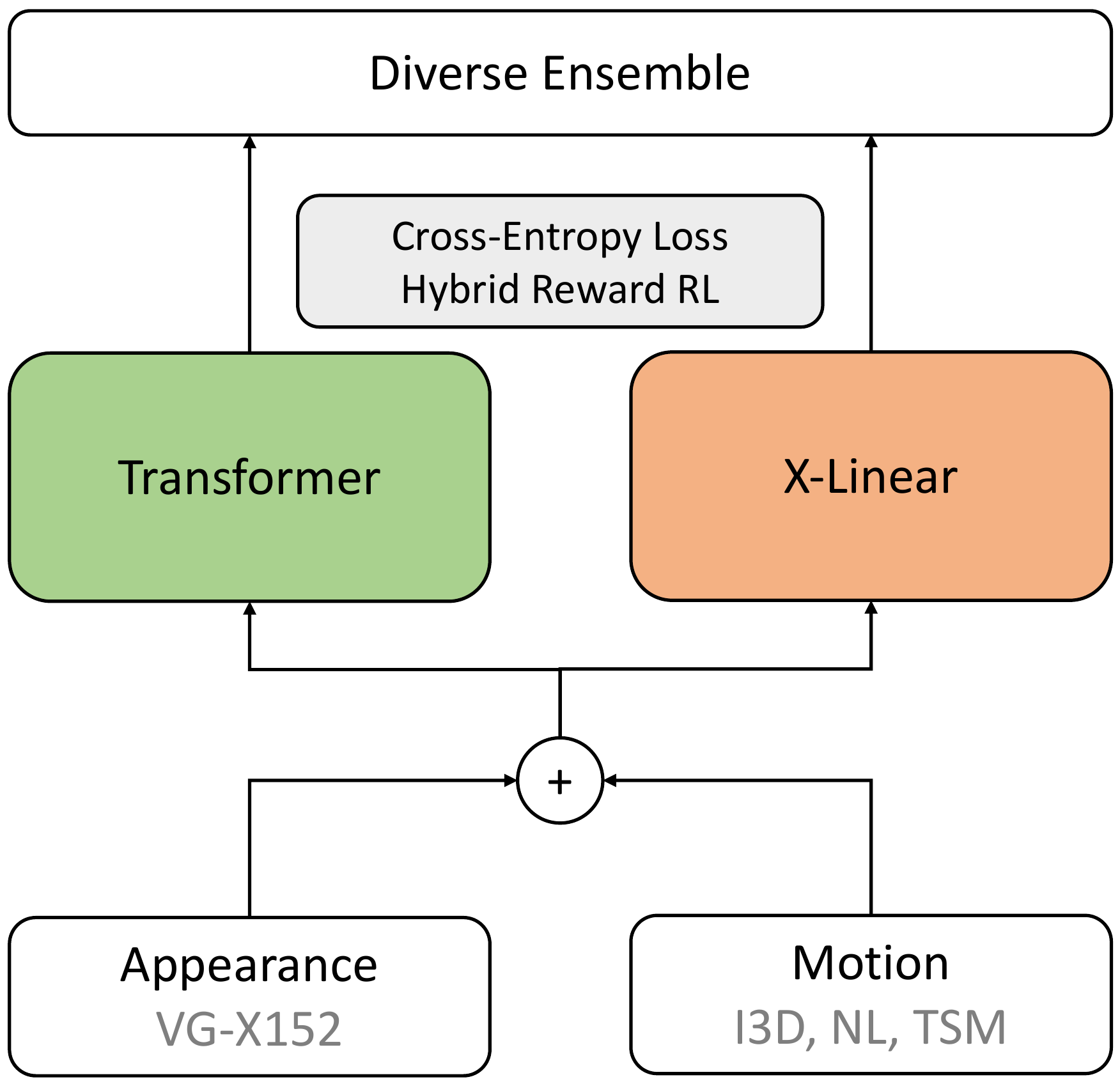}
	\end{center}
	\caption{The overall framework of our method.}
	\label{overall}
\end{figure*}

\section{Methods}
\paragraph{Multi-View Video Features}
To better extract the semantic features of the video, we refer to a variety of video feature extraction methods to extract more semantic features.
We hope to enhance the model's understanding of the different semantic information of the video by combining multiple features.
In this work, we mainly consider the motion and appearance features. 
Motion features operates mainly on the temporal dimension, which tells what activities the persons/objects in the video are involved in. 
 Our motion feature extraction networks include the I3D features \cite{carreira2017quo} provided by the vatex team; the nonlocal methods \cite{ghadiyaram2019large, tran2019video, tran2018closer}, and the TSM model \cite{lin2018temporal}. To better extract the video features, we use the strategy of randomly cropping video frames and randomly selecting partial videos. 
Appearance feature operates on the spatial dimension, which tells what objects are shown in the video. 
To fully capture the appearance of each key objects, we use an object detector to detect key objects in the key frame of each clip. 
Specifically, following \cite{anderson2018bottom}, we use Faster R-CNN \cite{Ren2017Faster} in conjunction with a stronger ResNeXt-152 backbone to extract object features for the key frames.  
The detector is pre-trained on the Visual Genome \cite{Krishna2017Visual} dataset for better feature representations. 
 The object features are mean-pooled to obtain the appearance feature for each key frame. 
 The resulting appearance feature in conjunction with the corresponding motion feature of that clip server as inputs to the captioning models.  

\paragraph{Model Architectures}
Compared to last year's challenge, we used a more powerful X-Linear \cite{pan2020x} model, which can achieve better performance than the top-down \cite{anderson2018bottom} model. X-Linear is an attention block that just came up in CVPR2020. The block can easily replace traditional attention block in LSTM or Transformer architecture.  X-Linear block exploits both the spatial and channel-wise bilinear attention distributions to capture the 2nd order interactions between the input single-modal or multi-modal features, selectively capitalizing on visual information or performing multimodal reasoning.

The Transformer \cite{vaswani2017attention} is another standard encoder-decoder framework, largely different from the lstm-based framework, and has advanced the state-of-the-art on various natural language processing tasks. It is a stack of multi-head attention substantially, using the self-attention mechanism to compute hidden representations of two arbitrary positional inputs. Disintegrate a whole video into a serial of clip features, video captioning task can be adjusted to fit the transformer framework. Inspirited by this insight, we transfer transformer architecture into video captioning tasks. 

\begin{table*}[h]
	\begin{center}
		\begin{tabular}{clccccccc}
			\toprule
			Language& Method & CIDEr & BLEU-1 & BLEU-2 & BLEU-3 & BLEU-4 & METEOR & ROUGE-L \\
			\midrule
			\multirow{3}*{Chinese}&VATEX-team \cite{wang2019vatex}         & 35.1 & 74.5 & 53.7 & 36.6 & 24.8 & 29.4 & 51.6\\
			&X-Linear & 56.8 & 81.6 & 63.5 & 45.9 & 32.2 & 31.9 & 56.1 \\
			&X-Linear+Transformer        & \textbf{59.5} & \textbf{82.2} & \textbf{64.3} & \textbf{46.5} & \textbf{32.6} & \textbf{32.1} & \textbf{56.5}\\
			
			\midrule
			\multirow{3}*{English}&VATEX-team \cite{wang2019vatex}         & 45.1 & 71.3 & 53.3 & 39.6 & 28.5 & 21.6 & 47.0\\
			&X-Linear & 76.3 & 81.9 & 66.5 & 52.1 & 39.4 & 25.2 & 53.0 \\
			&X-Linear+Transformer & \textbf{81.4} & \textbf{83.1} & \textbf{68.0} & \textbf{53.6} & \textbf{40.7} & \textbf{25.8} & \textbf{53.7}\\
			
			\bottomrule
		\end{tabular}
	\end{center}
	\caption{The ensemble results of our ultimate models on Vatex test set and \textbf{X-Linear+Transformer} is our final submission on the leaderboard. }
\end{table*}

\paragraph{Hybrid Reward For Reinforcement Learning}
Reinforcement Learning can further promote the performance of the model and it has been proved that the optimization for the CIDEr criterion can advance other criterions indirectly \cite{rennie2017self}. CIDEr, METEOR, ROUGE and BLEU  \cite{chen2015microsoft} are the criterions for the caption quality requested by the official VATEX website. The partial optimization for some criterions can result in better performance on the corresponding criterions. In this task, we find that using a hybrid reward, i.e. a linear combination of different metric scores, can result in a better overall result.
\paragraph{Diverse Ensemble of Models}
We adopt the common practice of ensemble methods, e.g. Average Ensemble and Weighted Ensemble. The experimental results demonstrate that Weighted Ensemble method has little improvement over Average Ensemble so we solely choose the former strategy. Our single models mainly belong to two different architectures, i.e. X-Linear and Transformer, and each type of model is reserved by four strategies for ensemble: the models with different initialized seeds; the models trained with different settings, i.e. the learning rate, scheduled sampling probability; the models with different visual features as inputs. Finally, we integrate all those models by average method.

\section{Experiments}
\subsection{Data pre-processing}
We train the models on the VATEX dataset following the official splits and preprocess similar procedures on all captions: both the English and the Chinese captions are truncated to a maximum of 30 words and transform all sentences to lowercase for the English caption. The word occurring less than 5 times in training captions is filtered and replaced by a special token UNK. We use the segmented Chinese words rather than raw Chinese characters. The vocabulary sizes of Chinese and English captions are 7,105 and 11,719 respectively.
\subsection{Implementation Details}

We conduct our experiments based on the Pytorch framework. We used a similar training strategy for training the English model and the Chinese model. We use Adam optimizer to train the model with a batch size of 64. Our training is divided into two phases, the first phase is cross entropy training and the second phase is reinforcement learning training. Label smoothing is set to 0.1. At the inference stage, we adopt the beam search strategy and set the beam size as 3. 
\paragraph{X-Linear Model}
For the X-Linear model setting, the dimensions of the bilinear query-key representation and the transformed bilinear feature in X-Linear attention block is set as 1,024 and 512, respectively. The hidden layer size in LSTM decoder is set as 1,024.  We stack four X-Linear attention blocks (plus ELU) in the encoder and the sentence decoder is equipped with one X-Linear attention block (plus ELU). The hidden layer size in LSTM decoder is set as 1,024. We set the maximum iteration as 70 epochs for the first learning stage since low-rank bilinear pooling may lead to slow convergence rate. And then we train the model with the learning rate of $1\times 10^{-5}$ for another 35 epochs

\paragraph{Transformer Model}
The setting in the Transformer model is similar to the X-Linear model. The latent dimensions in the self-attention and the feed-forwand network are 1024 and 4096. The number of heads $ h $ is 16 and the latent dimensions for each head $ d_{h} = d/h = 64 $. The number of attention blocks L in the encoder and decoder are 1. 
All the transformer models are trained as the following setting: the base learning rate is set to $\min(t \times 10^{-4}, 3 \times 10^{-4})$, where $t$ is the current epoch number that starts at 1. After 6 epochs, the learning rate is decayed by 1/2 every 3 epochs. The first training stage lasts for 15 epochs and the second training stage lasts for 10 epochs.

\section{Results}

In order to better improve the final performance, we used model ensemble strategy in the final testing stage to fuse all the models trained by different hyper-parameters. 
Since there are models that use data from the validation set, we only list the results of the test set.
We show the X-Linear results and X-Linear+Transformer results in the Table 1.
The final result of the VATEX Chinese combines the prediction results of 10 X-Linear models and 8 Transformer models.
At the same time, we show the results of the VATEX English in the official test set.
The final result combines 15 X-Linear models and 17 Transformer models.


\end{document}